\documentclass{article}
\usepackage{amsfonts,amsbsy}

\usepackage{amsmath}
\usepackage{amssymb}
\usepackage{enumerate}

\begin{document}
\pagestyle{empty}
{\sl{\small{\noindent Second International Workshop on\\ \noindent   Functional and Operatorial Statistics.\\ \noindent  Santander, June 16-18, 2011}}}\\~\\
\begin{center}{\Large{\bf Multiple functional regression with both discrete and continuous covariates}}
\end{center}
\begin{center}
\vspace{5mm}
{\bf{Hachem Kadri*$^{1}$, Philippe Preux$^{1,2}$, Emmanuel Duflos$^{1,3}$, St\'ephane Canu$^{4}$}}\\
\vspace{5mm}
\noindent $^1$SequeL Project, INRIA Lille - Nord Europe, Villeneuve d'Ascq, France\\
$^2$LIFL/CNRS, Universit\'e de Lille, Villeneuve d'Ascq, France\\
$^3$LAGIS/CNRS, Ecole Centrale de Lille, Villeneuve d'Ascq, France\\
$^4$LITIS, INSA de Rouen, St Etienne du Rouvray, France\\

\{hachem.kadri, philippe.preux, emmanuel.duflos\}@inria.fr, scanu@insa-rouen.fr

\vspace{5mm}
\rule{5cm}{.5mm}
\end{center}

\vspace{1cm}

\noindent {\bf\Large Abstract}\\

\vspace{-0.1cm}
In this paper we present a nonparametric method for extending functional regression methodology to the situation 
where more than one functional covariate is used to predict a functional response.
Borrowing the idea from Kadri et al.~(2010a), the method, which support mixed discrete and continuous explanatory variables, 
is based on estimating a function-valued function in reproducing kernel Hilbert spaces by virtue of positive 
operator-valued kernels. 

\vspace{0.8cm}

\noindent {\bf\Large 1. Introduction}\\

\vspace{-0.1cm}
The analysis of interaction effects between continuous variables in multiple regression has received a significant amount 
of attention from the research community. In recent years, a large part of research has been focused on functional regression 
where continuous data are represented by real-valued functions rather than by discrete, finite dimensional vectors. 
This is often the case in functional data analysis~(FDA) when observed data have been measured over a densely sampled grid. 
We refer the reader to Ramsay and Silverman~(2002, 2005) and Ferraty and Vieu~(2006) for more details on 
functional data analysis of densely sampled data or fully observed trajectories. In this context, various functional 
regression models~(Ramsay and Silverman, 2005) have been proposed according to the nature of explanatory~(or covariate) 
and response variables, perhaps the most widely studied is the generalized functional linear model where covariates are 
functions and responses are scalars~(Cardot et al., 1999,2003; James, 2002; M\"{u}ller and Stadtm\"{u}ller, 2005; 
Preda, 2007). 

In this paper, we are interested in the case of regression models with a functional response.~Two 
subcategories of such models have appeared in the FDA literature: covariates are scalars and responses are functions 
also known as \textquotedblleft functional response model\textquotedblright~(Faraway, 1997; Chiou et al., 2004); both covariates and responses are 
functions~(Ramsay and Dalzell, 1991; He et al., 2000; Cuevas et al., 2002; Prchal and Sarda, 2007; Antoch et al., 2008). 
In this work, we pay particular attention to this latter situation which corresponds to extending multivariate linear 
regression model to the functional case where all the components involved in the model are functions. 
Unlike most of previous works which consider only one functional covariate variable, we wish to perform a regression 
analysis in which multiple functional covariates are used to predict a functional response. The methodology which is 
concerned with solving such task is referred to as a multiple functional regression. 

Previous studies on multiple functional regression~(Han et al., 2007; Matsui et al., 2009; Valderrama et al., 2010) 
assume a linear relationship between functional 
covariates and responses and model this relationship via a multiple functional linear regression model which generalizes the 
model in Ramsay and Dalzell~(1991) to deal with more than one covariate variable. However, extensions to nonparametric 
models have not been considered. Nonparametric functional regression~(Ferraty and Vieu, 2002,2003) is addressed mostly in 
the context of functional covariates and scalar responses. More recently, Lian~(2007) and Kadri et al.~(2010a) showed how 
function-valued 
reproducing kernel Hilbert spaces~(RKHS) and operator-valued kernels can be used for the nonparametric estimation of the 
regression function when both covariates and responses are curves. Building on these works, we present in this paper a 
nonparametric multiple functional regression method where several functions would serve as predictors. Furthermore, we 
aim at extending this method to handle mixed discrete and functional explanatory variables. 
This should be helpful for situations where a subset of regressors are comprised of repeated observations of an outcome 
variable and the remaining are independent scalar or categorical variables. In Antoch et al.~(2008) for example, the authors 
 discuss the use of a functional linear regression model with a functional response to predict electricity consumption 
and mention that including the knowledge of special events such as festive days in the estimation procedure may improve the 
prediction.

The remainder of this paper is organized as follows. Section 2 reviews the multiple functional linear regression model and 
discusses its nonparametric extension. This section also describes the RKHS-based estimation procedure for the nonparametric 
multiple functional regression model. Section 3 concludes the paper. 

\vspace{0.8cm}

\noindent {\bf\Large 2. Multiple functional regression}\\

\vspace{-0.1cm}
Before presenting our nonparametric multiple function regression procedure, we start this section with a brief overview 
of the multiple functional linear regression model~(Matsui et al., 2009; Valderrama et al., 2010). 
This model extends functional linear regression with a functional 
response~(Ramsay and Dalzell, 1991; Ramsay and Silverman, 2005) to deal with more than one covariate and seeks to explain 
a functional response variable $y(t)$ by several functional covariates $x_k(s)$. A multiple functional linear regression 
model is formulated as follows:
\begin{equation}
\label{mflr}
 y_i(t) = \alpha(t) + \sum\limits_{k=1}^p\int_{I_s} x_{ik}(s)\beta_k(s,t) ds + \epsilon_i(t), 
\quad t\in I_t, \quad i=1,\ldots,n,
\end{equation}
where $\alpha(t)$ is the mean function, $p$ is the number of functional covariates, $n$ is the number of observations, 
$\beta_k(s,t)$ is the regression function for the $k$-th covariate and $\epsilon_i(t)$ a random error function. 
To estimate the functional parameters of this model, one can consider the centered covariate and response variables to 
eliminate the functional intercept $\alpha$. Then, $\beta_k(.,.)$ are approximated by a linear combination of basis 
functions and the corresponding real-valued basis coefficients can be estimated by minimizing a penalized least square 
criterion. Good candidates for the basis functions include the Fourier basis~(Ramsay and Silverman, 2005) and the B-spline 
basis~(Prchal and Sarda, 2007).

It is well known that parametric models suffer from the restriction that the input-output relationship has to be specified 
a priori. By allowing the data to model the relationships among variables, nonparametric models have emerged as a powerful 
approach for addressing this problem. In this context and from functional input-output data 
$(x_i(s),y_i(t))_{i=1}^n \in (\mathcal{G}_{x})^p \times \mathcal{G}_{y}$ where 
$\mathcal{G}_{x}:I_s\longrightarrow \mathbb{R}$ and $\mathcal{G}_{y}:I_t\longrightarrow \mathbb{R}$
, a nonparametric multiple functional regression model can be defined 
as follows:
\begin{equation}
\nonumber
\label{nmfr}
 y_i(t) = f(x_i(s)) + \epsilon_i(t), 
\quad s\in I_s,\ t\in I_t, \quad i=1,\ldots,n,
\end{equation}
where $f$ is a linear operator which perform the mapping between two spaces of functions. 
In this work, we consider a slightly modified model in which covariates could be a mixture of discrete and continuous 
variables. More precisely, we consider the following model
\begin{equation}
\label{mnmfr}
 y_i(t) = f(x_i) + \epsilon_i(t), 
\quad i=1,\ldots,n,
\end{equation}
where $x_i\in X$ is composed of two subsets $x_i^d$ and $x_i^c(s)$. $x_i^d\in \mathbb{R}^k$ is a $k\times 1$ vector of discrete 
dependent or independent variables and $x_i^c(s)$ is a vector of p continuous functions, so each $x_i$ contains $k$ discrete 
values and $p$ functional variables.

Our main interest in this paper is to design an efficient estimation procedure of the regression parameter $f$ of the 
model~(\ref{mnmfr}). An estimate $f^*$ of $f\in \mathcal{F}$ can be obtained by minimizing the following regularized 
empirical risk
\begin{equation}
\label{mp}
f^* = \arg\min\limits_{f \in \mathcal{F}}\sum\limits_{i=1}^{n}\|y_{i}-f(x_{i})\|_{\mathcal{G}_{y}}^{2}
+\lambda\|f\|_{\mathcal{F}}^{2}
\end{equation}
Borrowing the idea from Kadri et al.~(2010a), we use function-valued reproducing kernel Hilbert spaces~(RKHS) 
and operator-valued kernels to solve this minimization problem. 
Function-valued RKHS theory is the extension of the scalar-valued 
case to the functional response setting. In this context, Hilbert spaces of function-valued functions are constructed and basic 
properties of real RKHS are restated. Some examples of potential applications of these spaces can be found in 
Kadri et al.~(2010b) and in the area of multi-task learning (discrete outputs) see Evgeniou et al.~(2005). 
Function-valued RKHS theory is based on the \textit{one-to-one correspondence} between reproducing kernel Hilbert spaces of 
function-valued functions and positive operator-valued kernels. We start by recalling some basic properties of such Spaces. 
We say that a Hilbert space $\mathcal{F}$ of functions $ X\longrightarrow\mathcal{G}_{y}$ has the 
\textit{reproducing property}, if $\forall x\in X$ the evaluation functional $f\longrightarrow f(x)$ is continuous. This 
continuity is equivalent to the continuity of the mapping $f\longrightarrow \langle f(x),g\rangle_{\mathcal{G}_{y}}$ for any 
$x\in X$ and $g\in \mathcal{G}_{y}$.  By the Riesz representation theorem it follows that for a given $x\in X$ and for any 
choice of $g\in \mathcal{G}_{y}$, there exists an element 
$h_x^g \in \mathcal{F}$, s.t.
\begin{equation*}
 \forall f \in \mathcal{F}\ \ \ \langle h_x^g,f\rangle_{\mathcal{F}} = \langle f(x),g\rangle_{\mathcal{G}_{y}}
\end{equation*}
We can therefore define the corresponding operator-valued kernel $K(.,.)\in \mathcal{L(G}_{y})$, where 
$\mathcal{L(G}_{y})$ denote the set of bounded linear operators from $\mathcal{G}_{y}$ to  $\mathcal{G}_{y}$, such that
\begin{equation*}
 \langle K(x,z)g_1,g_2\rangle_{\mathcal{G}_{y}} = \langle h_x^{g_1},h_z^{g_2} \rangle_{\mathcal{F}}
\end{equation*}
It follows that $\langle h_x^{g_1}(z), g_2\rangle_{\mathcal{G}_{y}} = \langle h_x^{g_1},h_z^{g_2} \rangle_{\mathcal{F}} 
= \langle K(x,z)g_1,g_2\rangle_{\mathcal{G}_{y}}$ and thus we obtain the reproducing property
\begin{equation}
\nonumber
 \langle K(x,.)g,f\rangle_{\mathcal{F}} = \langle f(x),g\rangle_{\mathcal{G}_{y}} 
\end{equation}

It is easy to see that $K(x,z)$ is a positive kernel as defined below:\\
\textbf{Definition:} We say that $K(x,z)$, satisfying $K(x,z) = K(z,x)^*$, is a positive operator-valued kernel
if given an arbitrary finite set of points $\{(x_i,g_i)\}_{i=1,\ldots,n}\in X\times \mathcal{G}_{y}$, the corresponding 
block matrix $K$ with $K_{ij} = \langle K(x_i,x_j)g_i, g_j \rangle_{\mathcal{G}_{y}}$ is positive semi-definite. 

Importantly, the converse is also true. Any positive operator-valued kernel $K(x,z)$ gives rise
to an RKHS $\mathcal{F}_K$, which can be constructed by considering the space of function-valued functions $f$ having the 
form $f(.) = \sum_{i=1}^n K(x_i,.)g_i$ and taking completion with respect to the inner product given by 
$\langle K(x,.)g_1, K(z,.)g_2 \rangle_{\mathcal{F}} = \langle K(x,z)g_1,g_2 \rangle_{\mathcal{G}_{y}}$. 

The functional version of the Representer Theorem can be used to show that the solution of the minimization 
problem~(\ref{mp}) is of the following form:
\begin{equation}
 f^*(x) = \sum_{j=1}^n K(x,x_j)g_j
\end{equation}
Substituting this form in~(\ref{mp}) , we arrive at the following minimization over the scalar-valued functions $g_i$ 
rather than the function-valued function $f$ 
\begin{equation}
\label{mp1}
    \min\limits_{g\in (\mathcal{G}_{y})^n}
\sum\limits_{i=1}^{n}\|y_{i}-\sum\limits_{j=1}^{n}
K(x_{i},x_{j})g_{j}\|_{\mathcal{G}_{y}}^{2}  
     +\lambda
    \sum\limits_{i,j}^{n}\langle K(x_{i},x_{j})g_{i},g_{j}\rangle_{\mathcal{G}_{y}} 
\end{equation}
This problem can be solved by choosing a suitable operator-valued kernel. 
Choosing $K$ presents two major difficulties: we 
need to construct a function from an adequate operator, and which takes as arguments variables 
composed of scalars and functions. 
Lian~(2007) considered the identity operator, while in Kadri et al.~(2010) 
the authors showed that it will be more useful to choose other operators than identity that are able to take into account 
functional  properties of the input and output spaces. They also introduced a functional extension of the Gaussian kernel 
based on the multiplication operator. Using this operator, their approach can be seen as a nonlinear extension of the 
functional linear concurrent model~(Ramsay and Silverman, 2005). Motivated by extending the functional linear regression 
model with functional response, we consider in this work a kernel $K$ constructed from the integral operator and having 
the following form:
\begin{equation}
( K(x_i,x_j) g) (t)= [k_{x^d}(x_i^d, x_j^d) + k_{x^c}(x_i^c, x_j^c)]  \int k_y(s,t)g(s) ds
\end{equation}
where $k_{x^d}$ and $k_{x^c}$ are scalar-valued kernels on $\mathbb{R}^k$ and $(\mathcal{G}_{x})^p$ respectively and $k_y$ 
the reproducing kernel of the space $\mathcal{G}_{y}$. Choosing $k_{x^d}$ and $k_y$ is not a problem. Among the large number 
of possible classical kernels $k_{x^d}$ and $k_y$, we chose the Gaussian kernel. However, constructing $k_{x^c}$ is slightly 
more delicate. One can use the inner product in~$(\mathcal{G}_{x})^p$ to construct a linear kernel. Also, extending 
real-valued functional kernels such as those in Rossi et Villa.~(2006) to multiple functional inputs could be possible. 

To solve the problem~(\ref{mp1}), we consider that $\mathcal{G}_{y}$ is a real-valued RKHS and $k_y$ its reproducing kernel 
and then each function in this space can be approximated by a finite linear combination of kernels. So, the functions 
$g_i(.)$ can be approximated by $\sum_{l=1}^m \alpha_{il}k_y(t_l,.)$ and solving~(\ref{mp1}) returns to finding the 
corresponding real variables $\alpha_{il}$. Under this framework and using matrix formulation, we find that the 
$nm\times 1$ vector $\alpha$ satisfies the system of linear equation
\begin{equation}
 (\mathbf{K}+\lambda I) \alpha = Y
\end{equation}
where the $nm\times 1$ vector $Y$ is obtained by concatenating the columns of the matrix $(Y_{il})_{i\leq n ,\ l\leq m}$ and 
$\mathbf{K}$ is the block operator kernel matrix $(\mathbf{K}_{ij})_{1\leq i,j\leq n}$ where each 
$\mathbf{K}_{ij}$ is a $m\times m$ matrix.

\vspace{0.8cm}

\noindent {\bf\Large 3. Conclusion}\\

\vspace{-0.1cm}
We study the problem of multiple functional regression where several functional explanatory variables are used to predict a 
functional response. 
Using function-valued RKHS theory, we have proposed a nonparametric estimation procedure which support mixed discrete and 
continuous covariates. 
In future, we will illustrate our approach and evaluate its performance by experiments on simulated and real data.

\vspace{0.19cm}
\subsubsection*{Acknowledgments}
H.K. is supported by Junior Researcher Contract No.~4297 from the the Nord-Pas de Calais region.

\vspace{0.8cm}

\noindent {\bf\Large References}\\

\vspace{-0.1cm}
\small
Antoch, J., Prchal, L., De Rosa, M. and Sarda, P. (2008). Functional linear regression with functional response: 
application to prediction of electricity consumption. IWFOS 2008 Proceedings, Functional and operatorial statistics, 
Physica-Verlag, Springer.
%

\vspace{4.2pt}
Cardot, H., Ferraty, F. and Sarda, P. (1999). Functional linear model. Statistics and Probability Letters 45, 11-22.

\vspace{4.2pt}
Cardot, H., Ferraty, F.  and Sarda, P. (2003). Spline Estimators for the Functional Linear Model. Statistica Sinica, Vol. 13, 571-591. 

\vspace{4.2pt}
Chiou, J.M., M\"{u}ller, H.G. and Wang, J.L. (2004). Functional response models. Statistica Sinica 14, 675-693.

\vspace{4.2pt}
Cuevas, A., Febrero, M. and Fraiman, R. (2002). Linear functional regression: the case of fixed design and functional response. Can. J. Statist. 30, 285-300.

\vspace{4.2pt}
Evgeniou, T., Micchelli, C. A. and Pontil, M. (2005). Learning multiple tasks with kernel methods. 
Journal of Machine Learning Research, 6:615-637.

\vspace{4.2pt}
Faraway, J. (1997). Regression analysis for a functional response. Technometrics 39, 254-262.

\vspace{4.2pt}
Ferraty, F. and Vieu, P. (2002). The functional nonparametric model and applications to spectrometric data. 
Computational Statistics, 17, 545-564.

\vspace{4.2pt}
Ferraty, F. and Vieu, P. (2003). Curves discrimination: a nonparametric functional approach. 
Computational Statistics and Data Analysis, 44, 161-173.

\vspace{4.2pt}
Ferraty, F. and Vieu, P. (2006). Nonparametric Functional Data Analysis. N.Y. : Springer.

\vspace{4.2pt}
Han, S.W., Serban, N. and Rouse, B.W. (2007). Novel Perspectives On Market Valuation of Firms Via Functional Regression. 
Technical report, Statistics group, Georgia Tech.

\vspace{4.2pt}
He, G., M\"{u}ller, H.G. and Wang, J.L. (2000). Extending correlation and regression from multivariate to
functional data. Asymptotics in statistics and probability, Ed. Puri, M.L., VSP International Science
Publishers, 301-315.

\vspace{4.2pt}
James, G. (2002). Generalized linear models with functional predictors. J. Royal Statist. Soc. B 64, 411-432.

\vspace{4.2pt}
Kadri, H., Preux, P., Duflos, E., Canu, S. and Davy, M. (2010a). Nonlinear functional regression: a functional RKHS approach. 
in Proc. of the 13th Int'l Conf. on Artificial Intelligence and Statistics (AI \& Stats). JMLR: W\&CP 9, 374-380.

\vspace{4.2pt}
Kadri, H., Preux, P., Duflos, E., Canu, S. and Davy, M. (2010b). Function-Valued Reproducting Kernel Hilbert Spaces and 
Applications. NIPS workshop on TKML.

\vspace{4.2pt}
Lian, H. (2007). Nonlinear functional models for functional responses in reproducing kernel hilbert spaces. 
Canadian Journal of Statistics 35, 597-606.

\vspace{4.2pt}
Matsui, H., Kawano, S. and Konishi, S. (2009). Regularized functional regression modeling for functional response and 
predictors. Journal of Math-for-industry, Vol.1, 17-25.

\vspace{4.2pt}
M\"{u}ller, H.G. and Stadtm\"{u}ller, U. (2005). Generalized functional linear models. Annals of Statistics 33, 774-805.

\vspace{4.2pt}
Prchal, L. and Sarda, P. (2007). Spline estimator for the functional linear regression with functional response. Preprint.

\vspace{4.2pt}
Preda, C. (2007). Regression models for functional data by reproducing kernel Hilbert spaces methods. 
Journal of Statistical Planning and Inference, 137, 829-840.

\vspace{4.2pt}
Ramsay, J. and Dalzell, C.J. (1991). Some tools for functional data analysis. J. Royal Statist. Soc. Series B 53, 539-572.

\vspace{4.2pt}
Ramsay, J. and Silverman, B. (2002). Applied functional data analysis, New York: Springer.

\vspace{4.2pt}
Ramsay, J. and Silverman, B. (2005). Functional data analysis, New York: Springer.

\vspace{4.2pt}
Rossi, F. and Villa, N. (2006). Support vector machine for functional data classification. Neurocomputing, 69(7-9):730-742.

\vspace{4.2pt}
Valderrama, M.J., Oca\~na, F.A., Aguilera, A.M. and Oca\~na-Peinado, F.M. (2010). 
Forecasting pollen concentration by a two-Step functional model. Biometrics, 66:578-585. 
\end{document}